\documentclass[journal]{IEEEtran}
\usepackage[dvips]{graphicx}
\usepackage{epsfig,multirow}
\usepackage{amsmath,amssymb,bm}
\usepackage{amsfonts,dsfont,color,bbm}
\usepackage{algorithm,algorithmic,multicol,cite}
\usepackage{epsf,epstopdf}
\usepackage[font=footnotesize,labelfont=bf]{caption}
\usepackage{slashbox,subcaption}

\renewcommand{\vec}[1]{\mbox{\boldmath${#1}$}}
\newcommand{\norm}[1]{\left|\left|#1\right|\right|}
\newcommand{\nn}{\nonumber}

\newcommand{\Real}{{\mathbbm{R}}}
\newcommand{\Pre}{\mathcal{P}}
\newcommand{\R}{\mathcal{R}}
\newcommand{\F}{\mathcal{F}}
\newcommand{\Leaf}{\mathcal{L}}
\newcommand{\M}{\mathcal{M}}
\newcommand{\T}{\mathcal{T}}

\newcommand{\Set}{\mathcal{S}}

\newcommand{\sdt}{\left\{d[t] \right\}_{t \geq 1}}
\newcommand{\svxt}{\left\{\vec{x}[t] \right\}_{t \geq 1}}

\newcommand{\vx}{\vec{x}}
\newcommand{\va}{\vec{a}}
\newcommand{\vb}{\vec{b}}
\newcommand{\vw}{\vec{w}}
\newcommand{\vv}{\vec{v}}

\newcommand{\vp}{\vec{p}}

\newcommand{\vR}{\vec{R}}
\newcommand{\vI}{\vec{I}}

\newcommand{\vmu}{\vec{\mu}}

\title{Predicting Nearly As Well As the Optimal Twice Differentiable Regressor}

\author{N. Denizcan Vanli, Muhammed O. Sayin, and Suleyman S. Kozat, \textit{Senior Member, IEEE}
\thanks{This work was supported in part by the Turkish Academy of Sciences Outstanding Researcher Programme and in part by TUBITAK under Contract 112E161 and Contract 113E517.}
\thanks{The authors are with the Department of Electrical and Electronics Engineering,
Bilkent University, Bilkent, Ankara 06800, Turkey (e-mail: \{vanli,sayin,kozat\}@ee.bilkent.edu.tr).} }

\begin{document}
\maketitle

\begin{abstract}
We study nonlinear regression of real valued data in an individual sequence manner, where we provide results that are guaranteed to hold without any statistical assumptions. We address the convergence and undertraining issues of conventional nonlinear regression methods and introduce an algorithm that elegantly mitigates these issues via an incremental hierarchical structure, (i.e., via an incremental decision tree). Particularly, we present a piecewise linear (or nonlinear) regression algorithm that partitions the regressor space in a data driven manner and learns a linear model at each region. Unlike the conventional approaches, our algorithm gradually increases the number of disjoint partitions on the regressor space in a sequential manner according to the observed data. Through this data driven approach, our algorithm sequentially and asymptotically achieves the performance of the optimal twice differentiable regression function for any data sequence with an unknown and arbitrary length. The computational complexity of the introduced algorithm is only logarithmic in the data length under certain regularity conditions. We provide the explicit description of the algorithm and demonstrate the significant gains for the well-known benchmark real data sets and chaotic signals.
\end{abstract}


\section{Introduction}\label{sec:intro}

We study sequential nonlinear regression, where we aim to estimate or model a desired sequence $\sdt$ by using a sequence of regressor vectors $\svxt$. In particular, we seek to find the relationship, if it exists, between these two sequences, which is assumed to be unknown, nonlinear, and possibly time varying. This generic nonlinear regression framework is extensively studied in the machine learning and signal processing literatures since it can model a wide range of real life applications by capturing the salient characteristics of underlying signals and systems \cite{Linder1,Linder2,dc2,cart,ali,volterra,Hero,CTW,CTW2,add3,add7,Helmbold,drost,Takimoto1,Takimoto2}. In order to define and find this relationship between the desired sequence and regressor vectors, numerous methods such as neural networks, Volterra filters, and B-splines are used \cite{volterra,sayed_book,add1,add7,Linder1,Linder2,saf,fnf}. However, either these methods are extremely difficult to use in real life applications due to convergence issues, e.g., Volterra filters and B-splines, or it is quite hard to obtain a consistent performance in different scenarios, cf. \cite{volterra,Linder1,Linder2,dc1,dc2,Hero,CTW,RPTrees,KDTrees,Helmbold,drost,Takimoto1,Takimoto2}.

To this end, in this paper, we propose an algorithm that alleviates these issues by introducing hierarchical models that recursively and effectively partition the regressor space into subsequent regions in a data driven manner, where a different linear model is learned at each region. Unlike most of the nonlinear models, learning linear structures at each region can be efficiently managed. Hence, using this hierarchical piecewise model, we significantly mitigate the convergence and consistency issues. Furthermore, we prove that the resulting hierarchical piecewise model asymptotically achieves the performance of any twice differentiable regression function that is directly tuned to the underlying observations without any tuning of algorithmic parameters or without any assumptions on the data (other than an upper bound on the magnitude). Since most of the nonlinear modeling functions of the regression algorithms in the literature, such as neural networks and Volterra filters, can be accurately represented by twice differentiable functions \cite{volterra,Linder1,Linder2,saf,fnf,sayed_book}, our algorithm readily performs asymptotically as well as such nonlinear learning algorithms.

In particular, the introduced method sequentially and recursively divides the space of the regressors into disjoint regions according to the amount of the data in each region, instead of committing to a priori selected partition. In this sense, we avoid creating undertrained regions until a sufficient amount of data is observed. The nonlinear modeling power of the introduced algorithm is incremented (by consecutively partitioning the regressor space into smaller regions) as the observed data length increases. The introduced method adapts itself according to the observed data instead of relying on ad-hoc parameters that are set while initializing the algorithm. Thus, the introduced algorithm provides a significantly stronger modeling power with respect to the state-of-the-art methods in the literature as shown in our experiments.

We emphasize that piecewise linear regression using tree structures is extensively studied in the computational learning and signal processing literatures \cite{dc1,treelets,functional,RPTrees,KDTrees,Hero,CTW,CTW2,Helmbold,drost,Takimoto1,Takimoto2} due to its attractive convergence and consistency features. There exist several tree based algorithms that mitigate the overtraining problem by defining hierarchical piecewise models such as \cite{CTW,CTW2,Helmbold,drost,Takimoto1,Takimoto2}. Although these methods achieve the performance of the best piecewise model defined on a tree, i.e., the best pruning of a tree, they only yield satisfactory performance when the initial partitioning of the regressor space is highly accurate or tuned to the underlying data (which is unknown or even time-varying). Furthermore, there are more recent algorithms such as \cite{dc1} that achieve the performance of the optimal combination of all piecewise models defined on a tree that minimizes the accumulated loss. There are also methods that alleviate the overtraining problem by learning the region boundaries \cite{dc1} to minimize the regression error for a fixed depth tree with a computational complexity relatively greater compared to the ones in \cite{CTW,CTW2,Takimoto1,Takimoto2,Helmbold} (particularly, exponential in the depth of the tree). However, these algorithms can only provide a limited modeling power since the tree structure in these studies is fixed. Furthermore, the methods such as \cite{dc1} can only learn the locally optimal region boundaries due to the highly nonlinear (and non-convex) optimization structure. Unlike these methods, the introduced algorithm sequentially increases its nonlinear modeling power according to the observed data and directly achieves the performance of the best twice differentiable regression function that minimizes the accumulated regression error. We also show that in order to achieve the performance of a finer piecewise model defined on a tree, it is not even necessary to create these piecewise models when initializing the algorithm. Hence, we do not train a piecewise model until a sufficient amount of data is observed, and show that the introduced algorithm, in this manner, does not suffer any asymptotical performance degradation. Therefore, unlike the relevant studies in the literature, in which undertrained (i.e., unnecessary) partitions are kept in the overall structure, our method intrinsically eliminates the unnecessarily finer partitions without any loss in asymptotical performance (i.e., we maintain universality).

Aside from such piecewise linear regression techniques based on hierarchical models, there are various different methods to introduce nonlinearity such as B-splines and Volterra series \cite{volterra,Linder1,Linder2,saf,fnf,add1,add7}. In these methods, the nonlinearity is usually introduced by modifying the basis functions to create polynomial estimators, e.g., in \cite{fnf}, the authors use trigonometric functions as their basis functions. We emphasize that these techniques can be straightforwardly incorporated into our framework by using these methods at each region in the introduced algorithm to obtain piecewise nonlinear regressors. Note that the performance of such methods, e.g., B-splines and Volterra series (and other various methods with different basis functions), is satisfactory when the data is generated using the underlying basis functions of the regressor. In real life applications, the underlying model that generates the data is usually unknown. Thus, the successful implementation of these methods significantly depends on the match (or mismatch) between the regressor structure and the underlying model generating the data. On the other hand, the introduced algorithm achieves the performance of any such regressor provided that its basis functions are twice differentiable. In this sense, unlike the conventional methods in the literature, whose performances are highly dependant on the selection of the basis functions, our method can well approximate these basis functions (and regressors formed by these basis functions) via piecewise models such that the performance difference with respect to the best such regressor asymptotically goes to zero in a strong individual sequence manner without any statistical assumptions.

The main contributions of this paper are as follows. We introduce a sequential piecewise linear regression algorithm {\em i)} that provides a significantly improved modeling power by adaptively increasing the number of partitions according to the observed data, {\em ii)} that is highly efficient in terms of the computational complexity as well as the error performance, and {\em iii)} whose performance converges to {\em iii-a)} the performance of the optimal twice differentiable function that is selected in hindsight and {\em iii-b)} the best piecewise linear model defined on the incremental decision tree, with guaranteed upper bounds without any statistical or structural assumptions on the desired data as well as on the regressor vectors (other than an upper bound on them). Hence, unlike the state-of-the-art approaches whose performances usually depend on the initial construction of the tree, we introduce a method to construct a decision tree, whose depth (and structure) is adaptively incremented (and adjusted) in a data dependent manner, which we call an incremental decision tree. Furthermore, the introduced algorithm achieves this superior performance only with a computational complexity $O(\log(n))$ for any data length $n$, under certain regularity conditions. Even if these regularity conditions are not met, the introduced algorithm still achieves the performance of any twice differentiable regression function, however with a computational complexity linear in the data length.

The organization of the paper is as follows. We first describe the sequential piecewise linear regression problem in detail in Section \ref{sec:prob}. We then introduce the main algorithm in Section \ref{sec:algorithm} and prove that the performance of this algorithm is nearly as well as the best piecewise linear model that can be defined by the incremental decision tree in Section \ref{sec:proof}. Using this result, we also show that the introduced algorithm achieves the performance of the optimal twice differentiable function that is selected after observing the entire data before processing starts, i.e., non-causally. In Section \ref{sec:Simulations}, we demonstrate the performance of the introduced algorithm through simulations and then conclude the paper with several remarks in Section \ref{sec:Conclusion}.

\section{Problem Description}\label{sec:prob}
We study sequential nonlinear regression, where the aim is to estimate an unknown desired sequence $\sdt$ by using a sequence of regressor vectors $\svxt$, where the desired sequence and the regressor vectors are real valued and bounded but otherwise arbitrary, i.e., $d[t] \in \Real$, $\vx[t] \triangleq \left[ x_1[t],\dots,x_p[t] \right]^T \in \Real^p$ for an arbitrary integer $p$ and $|d[t]|, |x_i[t]| < A < \infty$ for all $t$ and $i=1,\dots,p$. We call the regressors as ``sequential'' if in order to estimate the desired data at time $t$, i.e., $d[t]$, they only use the past information $d[1],\dots,d[t-1]$ and the observed regressor vectors\footnote{All vectors are column vectors and denoted by boldface lower case letters. Matrices are denoted by boldface upper case letters. For a vector $\vx$, $\vx^T$ is the ordinary transpose. We denote $d_a^b \triangleq \{d[t]\}_{t=a}^b$.} $\vx[1],\dots,\vx[t]$.

In this framework, a piecewise linear model is constructed by dividing the regressor space into a union of disjoint regions, where in each region a linear model holds. As an example, suppose that the regressor space is parsed into $K$ disjoint regions $\R_1,\dots,\R_K$ such that $\bigcup_{k=1}^K \R_k = [-A, A]^p$. Given such a model, say model $m$, at each time $t$, the sequential linear\footnote{Note that affine models can also be represented as linear models by appending a $1$ to $\vx[t]$, where the dimension of the regressor space increases by one.} regressor predicts $d[t]$ as $\hat{d}_m[t] = \vv_{m,k}^T[t] \vx[t]$ when $\vx[t] \in \R_k$, where $\vv_{m,k}[t] \in \Real^p$ for all $k=1,\dots,K$. These linear models assigned to each region can be trained independently using different adaptive methods such as the least mean squares (LMS) or the recursive least squares (RLS) algorithms.

However, by directly partitioning the regressor space as $\bigcup_{k=1}^K \R_k = [-A, A]^p$ before the processing starts and optimizing only the internal parameters of the piecewise linear model, i.e., $\vv_{m,k}[t]$, one significantly limits the performance of the overall regressor since we do not have any prior knowledge on the underlying desired signal. Therefore, instead of committing to a single piecewise linear model with a fixed and given partitioning, and performing optimization only over the internal linear regression parameters of this regressor, one can use a decision tree to partition the regressor space and try to achieve the performance of the best partitioning over the whole doubly exponential number of different models represented by this tree \cite{double}.

\begin{figure}
  \centering
  \includegraphics[width=0.3\textwidth]{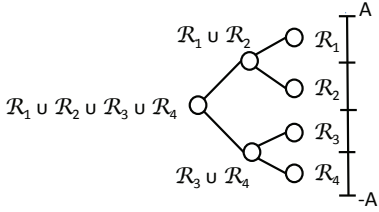}\\
  \caption{The partitioning of a one dimensional regressor space, i.e., $[-A,A]$, using a depth-$2$ full decision tree, where each node represents a portion of the regressor space.}\label{fig:tree}
\end{figure}

\begin{figure}
  \centering
  \includegraphics[width=0.48\textwidth]{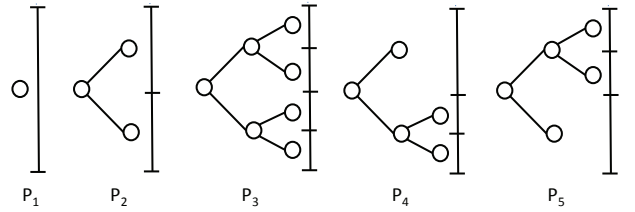}\\
  \caption{All different piecewise linear models that can be obtained using a depth-$2$ full decision tree, where the regressor space is one dimensional. These models are based on the partitioning shown in Fig. \ref{fig:tree}.}\label{fig:doublyexp}
\end{figure}

As an example, in Fig. \ref{fig:tree}, we partition the one dimensional regressor space $[-A,A]$, using a depth-$2$ tree, where the regions $\R_1,\dots,\R_4$ correspond to disjoint intervals on the real line and the internal nodes are constructed using union of these regions. In the generic case, for a depth-$d$ full decision tree, there exist $2^d$ leaf nodes and $2^d-1$ internal nodes. Each node of the tree represents a portion of the regressor space such that the union of the regions represented by the leaf nodes is equal to the entire regressor space $[-A,A]^p$. Moreover, the region corresponding to each internal node is constructed by the union of the regions of its children. In this sense, we obtain $2^{d+1}-1$ different nodes (regions) on the depth-$d$ decision tree (on the regressor space) and approximately $1.5^{2^d}$ different piecewise models that can be represented by certain collections of the regions represented by the nodes of the decision tree \cite{double}. For example, we consider the same scenario as in Fig. \ref{fig:tree}, where we partition the one dimensional real space using a depth-$2$ tree. Then, as shown in Fig. \ref{fig:tree}, there are $7$ different nodes on the depth-$2$ decision tree; and as shown in Fig. \ref{fig:doublyexp}, a depth-$2$ tree defines $5$ different piecewise partitions or models, where each of these models is constructed using certain unions of the nodes of the full depth decision tree.

We emphasize that given a decision tree of depth-$d$, the nonlinear modeling power of this tree is fixed and finite since there are only $2^{d+1}-1$ different regions (one for each node) and approximately $1.5^{2^d}$ different piecewise models (i.e., partitions) defined on this tree. Instead of introducing such a limitation, we recursively increment the depth of the decision tree as the data length increases. We call such a tree the ``incremental decision tree'' since the depth of the decision tree is incremented (and potentially goes to infinity) as the data length $n$ increases, hence in a certain sense, we can achieve the modeling power of an infinite depth tree. As shown in Theorem 2, the piecewise linear models defined on the tree will converge to any unknown underlying twice differentiable model under certain regularity conditions as $n$ increases.

To this end, we seek to find a sequential regression algorithm (whose estimate at time $t$ is represented by $\hat{d}_s[t]$), when applied to any sequence of data and regressor vectors, yields the following performance (i.e., regret) guarantee
\begin{equation}\label{eq:regret}
  \sum_{t=1}^n \left( d[t] - \hat{d}_s[t] \right)^2 - \inf_{f \in \F} \sum_{t=1}^n \left( d[t] - \hat{d}_f[t] \right)^2\leq o(n),
\end{equation}
over any $n$, without the knowledge of $n$, where $\F$ represents the class of all twice differentiable functions, whose parameters are set in hindsight, i.e., after observing the entire data before processing starts, and $\hat{d}_f[t]$ represents the estimate of the twice differentiable function $f\in\F$ at time $t$. The relative accumulated error in \eqref{eq:regret} represents the performance difference of the introduced algorithm and the optimal batch twice differentiable regressor. Hence, an upper bound of $o(n)$ in \eqref{eq:regret} implies that the algorithm $\hat{d}_s[t]$ sequentially and asymptotically converges to the performance of the regressor $\hat{d}_f[t]$, for any $f\in\F$.

\section{Nonlinear Regression via Incremental Decision Trees}\label{sec:algorithm}

In this section, we introduce the main results of the paper. Particularly, we first show that the introduced sequential piecewise linear regression algorithm asymptotically achieves the performance of the best piecewise linear model defined on the incremental decision tree (with possibly infinite depth) with the optimal regression parameters at each region that minimizes the accumulated loss. We then use this result to prove that the introduced algorithm asymptotically achieves the performance of any twice differentiable regression function. We provide the algorithmic details and the construction of the algorithm in Section \ref{sec:proof}.

{\bf Theorem 1:}
{\em Let $\sdt$ and $\svxt$ be arbitrary, bounded, and real-valued sequences of data and regressor vectors, respectively. Then the algorithm $\hat{d}[t]$ (given in Fig. \ref{alg}) when applied to these data sequences yields
\begin{align}%\label{eq:theorem}
  &\sum_{t=1}^n \left( d[t] - \hat{d}[t] \right)^2 \nn\\
  &\hspace{.5cm} - \inf_{m \in \M_n'} \left[ \inf_{\substack{\vv_{m,k} \in \Real^p \\ k=1,\dots,K_m}} \left\{ \sum_{t=1}^n \left( d[t] - \hat{d}_b[t] \right)^2 + \delta \norm{\vv_m}^2 \right\} \right] \nn\\
  &\hspace{1cm} \leq O\left( p \log^2(n) \right), \nn
\end{align}
for any $n$, with a computational complexity upper bounded by $O(n)$, where $K_m$ denotes the number of leaf nodes in the hierarchical model $m$, $\M_n$ represents the set of all hierarchical models defined on the incremental decision tree at time $n$, $\M_n'$ represents the set of all hierarchical models with at most $O(\log(n))$ leaves defined on the incremental decision tree at time $n$, i.e., $\M_n' \triangleq \{m \in \M_n : K_m \leq O(\log(n))\}$, and $\vv_m \triangleq [\vv_{m,1};\dots;\vv_{m,K_m}]$.
} \\

This theorem indicates that the introduced algorithm can asymptotically and sequentially achieve the performance of any piecewise model in the set $\M_n'$, i.e., the piecewise models having at most $O(\log(n))$ leaves defined on the tree. In particular, over any unknown data length $n$, the performance of the piecewise models with $O(\log(n))$ leaves can be sequentially achieved by the introduced algorithm with a regret upper bounded by $O\left( p \log^2(n) \right)$. In this sense, we do not compare the performance of the introduced algorithm with a fixed class of regressors, over any data length $n$. Instead, the regret of the introduced algorithm is defined with respect to a set of piecewise linear regressors, whose number of partitions are upper bounded by $O(\log(n))$, i.e., the competition class grows as $n$ increases. In the conventional tree based regression methods, the depth of the tree is set before processing starts and the performance of the regressor is highly sensitive with respect to the unknown data length. For example, if the depth of the tree is large whereas there are not enough data samples, then the piecewise model will be undertrained and yield an unsatisfactory performance. Similarly, if the depth of the tree is small whereas huge number of data samples are available, then trees (and regressors) with higher depths (and finer regions) can be better trained. As shown in Theorem 1, the introduced algorithm elegantly and intrinsically makes such decisions and performs asymptotically as well as any piecewise regressor in the competition class that grows exponentially in $n$ \cite{double}. Such a significant performance is achieved with a computational complexity upper bounded by $O(n)$, i.e., only linear in the data length, whereas the number of different piecewise models defined on the incremental decision tree can be in the order of $1.5^n$ \cite{double}. Moreover, under certain regularity conditions the computational complexity of the algorithm is $O(\log(n))$ as will be discussed in Remark 2. This theorem is an intermediate step to show that the introduced algorithm yields the desired performance guarantee in \eqref{eq:regret}, and will be used to prove the next theorem.

Using Theorem 1, we introduce another theorem presenting the main result of the paper, where we define the performance of the introduced algorithm with respect to the class of twice differentiable functions as in \eqref{eq:regret}.

{\bf Theorem 2:}
{\em Let $\sdt$ and $\svxt$ be arbitrary, bounded, and real-valued sequences of data and regressor vectors, respectively. Let $\F$ be the class of all twice differentiable functions such that for any $f \in \F$, $\frac{\partial^2 f(\vx)}{\partial x_i \partial x_j} \leq D < \infty$, $i,j=1,\dots,p$ and we denote $\hat{d}_f[t] = f(\vx[t])$. Then the algorithm $\hat{d}[t]$ given in Fig. \ref{alg} when applied to these data sequences yields
\begin{equation}%\label{eq:corollary}
  \sum_{t=1}^n \left( d[t] - \hat{d}[t] \right)^2 - \inf_{f \in \F} \sum_{t=1}^n \left( d[t] - \hat{d}_f[t] \right)^2 \leq o(p^2n), \nn
\end{equation}
for any $n$, with a computational complexity upper bounded by $O(n)$.
} \\

This theorem presents the nonlinear modeling power of the introduced algorithm. Specifically, it states that the introduced algorithm can asymptotically achieve the performance of the optimal twice differentiable function that is selected after observing the entire data in hindsight. Note that there are several kernel and neural network based sequential nonlinear regression algorithms \cite{volterra,Linder1,Linder2} (which can be modeled via twice differentiable functions) whose computational complexities are similar to the introduced algorithm. However, the performances of such nonlinear models are only comparable with respect to their batch variants. On the other hand, we demonstrate the performance of the introduced algorithm with respect to a extremely large class of regressors without any statistical assumptions. In this sense, the performance of any regression algorithm that can be modeled by twice differentiable functions is asymptotically achievable by the introduced algorithm. Hence, the introduced algorithm yields a significantly more robust performance with respect to the such conventional approaches in the literature as also illustrated in different experiments in Section \ref{sec:Simulations}.

The proofs of Theorem 1, Theorem 2, and the construction of the algorithm are given in the following section.

\section{Construction of the Algorithm and Proofs of the Theorems}\label{sec:proof}
In this section, we first introduce a labeling to efficiently manage the hierarchical models and then describe the algorithm in its main lines. We next prove Theorem 1, where we also provide the complete construction of the algorithm. We then present a proof for Theorem 2, using the results of Theorem 1.

\subsection{Notation}
We first introduce a labeling for the tree nodes following \cite{Willems}. The root node is labeled with an empty binary string $\lambda$ and assuming that a node has a label $\kappa$, where $\kappa = \nu_1 \dots \nu_l$ is a binary string of length $l$ formed from letters $\nu_1,\dots,\nu_l$, we label its upper and lower children as $\kappa1$ and $\kappa0$, respectively. Here, we emphasize that a string can only take its letters from the binary alphabet, i.e., $\nu \in \{0,1\}$, where $0$ refers to the lower child, and $1$ refers to the upper child of a node. We also introduce another concept, i.e., the definition of the prefix of a string. We say that a string $\kappa' = \nu'_{1} \dots \nu'_{l'}$ is a prefix to string $\kappa = \nu_1 \dots \nu_l$ if $l' \leq l$ and $\nu'_i = \nu_i$ for all $i=1,\dots,l'$, and the empty string $\lambda$ is a prefix to all strings. Finally, we let $\Pre(\kappa)$ represent all prefixes to the string $\kappa$, i.e., $\Pre(\kappa) \triangleq \{ \kappa_0,\dots,\kappa_l \}$, where $l \triangleq l(\kappa)$ is the length of the string $\kappa$, $\kappa_i$ is the string with $l(\kappa_i) = i$, and $\kappa_0 = \lambda$ is the empty string, such that the first $i$ letters of the string $\kappa$ forms the string $\kappa_i$ for $i=0,\dots,l$. Letting $\Leaf$ denote the set of leaf nodes for a given decision tree, each leaf node of the tree, i.e., $\kappa \in \Leaf$, is given a specific index $\alpha_\kappa \in \{0,\dots,M-1\}$ representing the number of regressor vectors that has fallen into $\R_\kappa$. For presentation purposes, we consider $M=2$ throughout the paper.

\subsection{Outline of the Algorithm}
At time $t=0$, the introduced algorithm starts with a single node (i.e., the root node) representing the entire regressor space. As the new data is observed, the proposed algorithm sequentially divides the regressor space into smaller disjoint regions according to the observed regressor vectors. In particular, each region is divided into subsequent child regions as soon as a new regressor vector has fallen into that region. In this incremental hierarchical structure, we assign an independent linear regressor to each node (i.e., to each region). Such a hierarchical structure (embedded with linear regressors) can define $1.5^n$ different piecewise linear models or partitions. We then combine the outputs of all these different piecewise models via a mixture of experts approach instead of committing to a single model. However, even for a small $n$, the number of piecewise models (i.e., experts) grows extremely rapidly (particularly, exponential in $n$). Hence, in order to perform this calculation in an efficient manner, we assign a weight to each node on the tree and present a method to calculate the final output using these weights with a significantly reduced computational complexity, i.e., logarithmic in $n$ under certain regularity conditions.

We then compare the performance of the introduced algorithm with respect to the best batch piecewise model defined on the incremental decision tree. Our algorithm first suffers a ``constructional regret'' that arise from the adaptive construction of the incremental decision tree (since the finer piecewise models are not present at the beginning of the processing) and from the sequential combination of the outputs of all piecewise models (i.e., due to the mixture of experts approach). Second, each piecewise model suffers a ``parameter regret'' while sequentially learning the true regression parameters at each region. We provide deterministic upper bounds on these regrets and illustrate that the introduced algorithm is twice-universal, i.e., universal in both entire piecewise models (even though the finer models appear as $n$ increases and do not used until then) and linear regression parameters.

\subsection{Proof of Theorem 1 and Construction of the Algorithm}
In this section, we describe the algorithm in detail and derive a regret upper bound with respect to the best batch piecewise model defined on the incremental decision tree.

Before the processing starts, i.e., at time $t=0$, we begin with a single node, i.e., the root node $\lambda$, having index $\alpha_\lambda=0$. Then, we recursively construct the decision tree according to the following principle. For every time instant $t>0$, we find the leaf node of the tree $\kappa \in \Leaf$ such that $\vx[t] \in \R_\kappa$. For this node, if we have $\alpha_\kappa=0$, we do not modify the tree but only increment this index by $1$. On the other hand, if $\alpha_\kappa=1$, then we generate two children nodes $\kappa0,\kappa1$ for this node by dividing the region $\R_\kappa$ into two disjoint regions $\R_{\kappa0}, \R_{\kappa1}$, using the plane $x_i = c$, where $i-1 \equiv l(\kappa) \pmod{p}$ and $c$ is the midpoint of the region $\R_\kappa$ along the $i$th dimension. For node $\kappa\nu$ with $\vx[t] \in \R_{\kappa\nu}$ (i.e., the children node containing the current regressor vector), we set $\alpha_{\kappa\nu}=1$ and the index of the other child is set to $0$. The accumulated regressor vectors and the data in node $\kappa$ are also transferred to its children to train a linear regressor in these child nodes.

\begin{figure}
  \centering
  \includegraphics[width=0.4\textwidth]{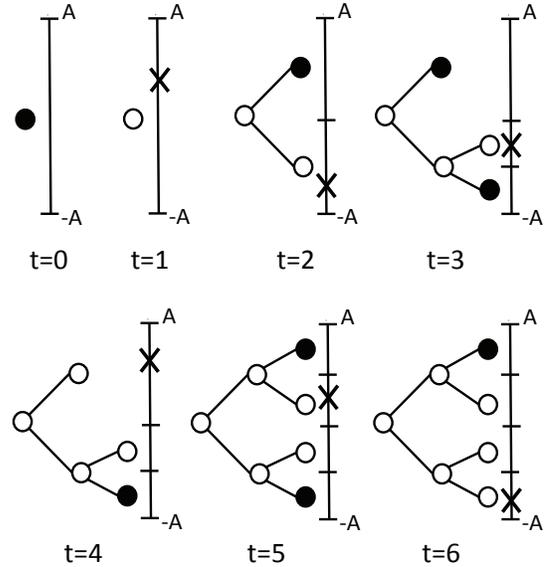}\\
  \caption{A sample evolution of the incremental decision tree, where the regressor space is one dimensional. The ``$\times$'' marks on the regressor space represents the value of the regressor vector at that specific time instant. Light nodes are the ones having an index of $1$, whereas the index of the dark nodes is $0$.}\label{fig:evolution}
\end{figure}

As an example, in Fig. \ref{fig:evolution}, we consider that the regressor space is one dimensional, i.e., $[-A,A]$, and present a sample evolution of the tree. In the figure, the nodes having an index of $0$ are shown as dark nodes, whereas the others are light nodes, and the regressor vectors are marked with $\times$'s in the one dimensional regressor space. For instance at time $t=2$, we have a depth-$1$ tree, where we have two nodes $0$ and $1$ with corresponding regions $\R_0 = [-A,0]$, $\R_1 = [0,A]$, and $\alpha_0 = 1$, $\alpha_1 = 0$. Then, at time $t=3$, we observe a regressor vector $\vx[3] \in \R_{0}$ and divide this region into two disjoint regions using $x_1 = -A/2$ line. We then find that $\vx[3] \in \R_{01}$, hence set $\alpha_{01} = 1$, whereas $\alpha_{00} = 0$.

We assign an independent linear regressor to each node on the incremental decision tree. Each linear regressor is trained using only the information contained in its corresponding node. Hence, we can obtain different piecewise models by using a certain collection of this node regressors according to the hierarchical structure. Each such piecewise model suffers a parameter regret in order to sequentially learn the optimal linear regression parameters at each region that minimizes the cumulative error. This issue is discussed towards the end of this section.

Using this incremental hierarchical structure with linear regressors at each region, the incremental decision tree can represent up to $1.5^n$ different piecewise linear models after observing data of length $n$. For example, in Fig. \ref{fig:evolution}, at time $t=6$, we have $5$ different piecewise linear models (see Fig. \ref{fig:doublyexp}), whereas at time $t=4$, we have $3$ different piecewise linear models. Each of these piecewise linear models can be used to perform the estimation task. However, we use a mixture of experts approach to combine the outputs of all piecewise linear models, instead of choosing a single one among them.

To this end, one can assign a performance dependent weight to each piecewise linear model defined on the incremental decision tree and combine their weighted outputs to obtain the final estimate of the algorithm \cite{kozat,convex,sayed_book}. In a conventional setting, such a mixture of expert approach is guaranteed to asymptotically achieve the performance of the best piecewise linear model defined on the tree \cite{kozat,convex,sayed_book}. However, in our framework, to achieve the performance of the best twice differentiable regression function, as $t$ increases (i.e., we observe new data), the total number of different piecewise linear models can increase exponentially with $t$. In this sense, we have a highly dynamic optimization framework. For example, in Fig. \ref{fig:evolution}, at time $t=4$, we have $3$ different piecewise linear models, hence calculate the final output of our algorithm as $\hat{d}[t] = w_1[t]\hat{d}_1[t] +  w_2[t]\hat{d}_2[t] + w_2[t]\hat{d}_2[t]$, where $\hat{d}_i[t]$ represents the output of the $i$th piecewise linear model and $w_i[t]$ represents its weight. However, at time $t=6$, we have $5$ different piecewise linear models, i.e., $\hat{d}[t] = \sum_{i=1}^5 w_i[t]\hat{d}_i[t]$, therefore the number of experts increases. Hence, not only such a combination approach requires the processing of the entire observed data at each time $t$ (i.e., it results in a brute-force batch-to-online conversion), but also it cannot be practically implemented even for a considerably short data sequences such as $n=100$.

To elegantly solve this problem, we assign a weight to each node on the incremental decision tree, instead of using a conventional mixture of experts approach. In this way, we illustrate a method to calculate the original highly dynamic combination weights in an efficient manner, i.e., without requiring the processing of the entire data for each new sample, and with a significantly reduced computational complexity.

To accomplish this, to each leaf node $\kappa \in \Leaf$, we assign a performance dependant weight \cite{Willems} as follows
\[
P_\kappa(n) \triangleq \exp\left\{ -\frac{1}{2a} \sum_{t \leq n \, : \, \vx[t] \in \R_\kappa} \left( d[t] - \hat{d}_{m,k}[t] \right)^2 \right\},
\]
where $\hat{d}_{m,k}[t]$ represents the linear regressor assigned to the $k$th node of the $m$th piecewise model and is constructed using the regressor introduced in \cite{Singer2} and discussed in \eqref{eq:first}. Then, we define the weight of an inner node $\kappa \notin \Leaf$ as follows \cite{Willems}
\begin{align}
  P_\kappa(n) &\triangleq \frac{1}{2} P_{\kappa0}(n) P_{\kappa1}(n) \nn\\
    &\hspace{.5cm} + \frac{1}{2} \exp\left\{ -\frac{1}{2a} \sum_{t \leq n \, : \, \vx[t] \in \R_\kappa} \left( d[t] - \hat{d}_{m,k}[t] \right)^2 \right\}. \nn
\end{align}
Using this definitions, the weight of the root node $\lambda$ can be constructed as follows
\begin{equation}
  P_\lambda(n) = \sum_{m \in \M_n} 2^{-B_m} P(n | m), \nn
\end{equation}
where
\begin{equation}
  P(n | m) \triangleq \exp\left\{ -\frac{1}{2a} \sum_{t=1}^n \left( d[t] - \hat{d}_m[t] \right)^2 \right\} \nn
\end{equation}
represents the performance of a given partition $m \in \M_n$ over a data length of $n$, and $B_m$ represents the number of bits required to represent the model $m$ on the binary tree using a universal code \cite{coding}.

Hence, the performance of the root node satisfies $P_\lambda(n) \geq 2^{-B_m} P(n|m)$ for any $m \in \M_n$. That is,
\begin{align}\label{eq:second}
  -2a \ln\left( P_\lambda(n) \right) &\leq \min_{m \in \M_n} \left\{ \sum_{t=1}^n \left( d[t] - \hat{d}_m[t] \right)^2 \right\} \nn\\
    & \hspace{.5cm} + 2a\ln(2)\log(n) + 4A^2 K_m \log(n),
\end{align}
where the last line follows when we maximize $B_m$ with respect to $m \in \M_n$ and the regret term $4A^2 K_m \log(n)$ follows due to the adaptive construction of the incremental decision tree. This upper bound corresponds to the constructional regret of our algorithm.

Hence, we have obtained a weighting assignment achieving the performance of the optimal piecewise linear model. We next introduce a sequential algorithm achieving $P_\lambda(n)$. To this end, we first note that we have
\begin{equation}
  P_\lambda(n) = \prod_{t=1}^n \frac{P_\lambda(t)}{P_\lambda(t-1)}.
\end{equation}
Now if we can demonstrate a sequential algorithm whose performance is greater than or equal to $P_\lambda(t) / P_\lambda(t-1)$ for all $t$, we can conclude the proof. To this end, we present a sequential update from $P_\lambda(t-1)$ to $P_\lambda(t)$.

After the structural updates, i.e., the growth of the incremental decision tree, are completed, say at time $t$, we observe a regressor vector $\vx[t] \in \R_\kappa$ for some $\kappa \in \Leaf$. Then, we can compactly denote the weight of the root node at time $t-1$ as follows
\begin{equation}
  P_\lambda(t-1) = \hspace{-.3cm} \sum_{\kappa_i \in \Pre(\kappa)} \hspace{-.3cm} \pi_{\kappa_i}[t-1] \exp\Bigg\{ -\frac{1}{2a} \hspace{-.3cm} \sum_{\substack{t' < t \\ \vx[t'] \in \R_{\kappa_i}}} \hspace{-.3cm} \left( d[t'] - \hat{d}_{\kappa_i}[t'] \right)^2 \Bigg\}, \nn
\end{equation}
where $\hat{d}_{\kappa}[t]$ represents the output of the regressor for node $\kappa$, $\kappa_i \in \Pre(\kappa)$ is the string formed from the first $i$ letters of $\kappa = \nu_1 \dots \nu_l$, and $\pi_{\kappa_i}[t]$ is recursively defined as follows
\[
\pi_{\kappa_i}[t] \triangleq
\begin{cases}
  \frac{1}{2}  &\mbox{, if $i=0$} \\
  \frac{1}{2} P_{\kappa_{i-1}\nu_i^c}(t-1) \pi_{\kappa_{i-1}}[t]  &\mbox{, if $1 \leq i \leq l-1$} \\
  P_{\kappa_{i-1}\nu_i^c}(t-1) \pi_{\kappa_{i-1}}[t]  &\mbox{, if $i=l$}
\end{cases}.
\]

Since $\vx[t] \in \R_\kappa$ for some $\kappa \in \Leaf$, then after $d[t]$ is revealed, the weight of the root node at time $t$ can be calculated as follows
\begin{align}
  P_\lambda(t) & = \sum_{\kappa_i \in \Pre(\kappa)} \pi_{\kappa_i}[t-1] \exp\left\{ -\frac{1}{2a} \left( d[t] - \hat{d}_{\kappa_i}[t] \right)^2 \right\} \nn\\
   & \hspace{0.5cm} \times \exp\Bigg\{ -\frac{1}{2a} \sum_{t' < t \, : \, \vx[t'] \in \R_{\kappa_i}} \left( d[t'] - \hat{d}_{\kappa_i}[t'] \right)^2 \Bigg\}, \nn
\end{align}
which results in
\begin{equation}\label{eq:conditional}
  \frac{P_\lambda(t)}{P_\lambda(t-1)} = \sum_{\kappa_i \in \Pre(\kappa)} \mu_{\kappa_i}[t-1] \exp\left\{ -\frac{1}{2a} \left( d[t] - \hat{d}_{\kappa_i}[t] \right)^2 \right\},
\end{equation}
where
\[
\mu_{\kappa_i}[t-1] \triangleq \frac{\pi_{\kappa_i}[t-1] \exp\left\{ -\frac{1}{2a} \sum_{\hspace{-0.3cm}\substack{t' < t \\ \vx[t'] \in \R_{\kappa_i}}} \hspace{-0.4cm} \left( d[t'] - \hat{d}_{\kappa_i}[t'] \right)^2 \right\}}{P_\lambda(t-1)}.
\]

We then focus on \eqref{eq:conditional} and observe that we have $\sum_{\kappa_i \in \Pre(\kappa)} \mu_{\kappa_i}[t-1] = 1$, which means that if the second term in \eqref{eq:conditional}, i.e.,
\begin{equation}
  f(\hat{d}_{\kappa_i}[t]) \triangleq \exp\left\{ -\frac{1}{2a} \left( d[t] - \hat{d}_{\kappa_i}[t] \right)^2 \right\}, \nn
\end{equation}
is concave, then by Jensen's inequality, we can conclude that
\begin{equation}\label{eq:sequential}
  \exp\left\{ -\frac{1}{2a} \left( d[t] - \hspace{-0.3cm} \sum_{\kappa_i \in \Pre(\kappa)} \hspace{-0.2cm} \mu_{\kappa_i}[t-1] \hat{d}_{\kappa_i}[t] \right)^2 \right\} \geq P_\lambda(t \, | \, t-1).
\end{equation}
Since the function $f(\hat{d}_{\kappa_i}[t])$ is concave when $\left( d[t] - \hat{d}_{\kappa_i}[t] \right)^2 < a$, and we have $|d[t]| \leq A$, we have to set $a \geq 4A^2$. Therefore, we obtain a sequential regressor in \eqref{eq:sequential}, whose performance is greater than or equal to the performance of the root node, and the final estimate of our algorithm is calculated as follows
\begin{equation}\label{eq:final_estimate}
  \hat{d}[t] \triangleq \sum_{\kappa_i \in \Pre(\kappa)} \mu_{\kappa_i}[t-1] \hat{d}_{\kappa_i}[t].
\end{equation}

Hence, our algorithm can achieve the performance of the best piecewise linear model defined on the incremental tree with a constructional regret given in \eqref{eq:second}. In order to achieve the performance of the best ``batch'' piecewise linear model, the introduced algorithm also suffers a parameter regret while learning the true regression parameters at each region. An upper bound on this regret is calculated as follows.

Consider an arbitrary piecewise model defined on the incremental decision tree, say the $m$th model, having $K_m$ disjoint regions $\R_1,\dots,\R_{K_m}$ such that $\bigcup_{k=1}^{K_m} \R_k = [-A, A]^p$. Then, a piecewise linear regressor can be constructed using the universal linear predictor of \cite{Singer2} in each region as $\hat{d}_m[t] = \vv_{m,k}^T[t] \, \vx[t]$, when $\vx[t] \in \R_k$, with the regression parameters $\vv_{m,k}[t] = \left( \vR_k[t] + \delta \vI \right)^{-1} \vp_k[t]$, where $\vI$ represents the appropriate sized identity matrix, $\vR_k[t] \triangleq \sum_{t' \leq t \, : \, \vx[t'] \in \R_k} \vx[t'] \, \vx^T[t']$, and $\vp_k[t] \triangleq \sum_{t' < t \, : \, \vx[t'] \in \R_k} d[t'] \, \vx[t']$. The upper bound on the performance of this regressor can be calculated following similar lines to \cite{Singer2} and it is obtained as follows
\begin{align}\label{eq:first}
  \sum_{t=1}^n \left( d[t] - \hat{d}_m[t] \right)^2 & - \min_{\substack{\vv_{m,k} \in \Real^p \\ k=1,\dots,K_m}} \bigg\{ \sum_{t=1}^n \left( d[t] - \hat{d}_b[t] \right)^2 \nn\\
  & \hspace{-1.3cm} + \delta \norm{\vv_m}^2 \bigg\} \leq  A^2 K_m p \ln\left( n/K_m \right) + O(1).
\end{align}
We emphasize that in each region of a piecewise model, different learning algorithms, e.g., different linear regressors or nonlinear ones, from the broad literature can be used. Note that although the main contribution of the paper is the hierarchical organization and efficient management of these piecewise models, we also discuss the implementation of a piecewise linear model \cite{Singer2} into our framework for completeness.

Finally, we achieve an upper bound on the performance of the introduced algorithm with respect to the best batch piecewise linear model. Combining the results in \eqref{eq:second} and \eqref{eq:first}, we obtain
\begin{align}
  \sum_{t=1}^n \left( d[t] - \hat{d}[t] \right)^2 & \leq \min_{m \in \M_n} \left\{ \sum_{t=1}^n \left( d[t] - \hat{d}_m[t] \right)^2 \right\} \nn\\
    & \hspace{.5cm} + 2a\ln(2)\log(n) + 4 A^2 K_m \log(n) \nn\\
    & \hspace{-2.7cm} \leq \min_{m \in \M_n} \left[ \min_{\substack{\vv_{m,k} \in \Real^p \\ k=1,\dots,K_m}} \left\{ \sum_{t=1}^n \left( d[t] - \hat{d}_b[t] \right)^2 + \delta \norm{\vv_m}^2 \right\} \right] \nn\\
    & \hspace{-2.6cm} + \underbrace{A^2 K_m \left( p \ln\left( n/K_m \right) + 4\log(n) \right) \hspace{-.05cm} + \hspace{-.05cm} 2a\ln(2)\log(n) \hspace{-.05cm} + \hspace{-.05cm} O(1)}_{\leq O(p \log^2(n))} , \nn
\end{align}
where the upper bound on the regret follows when $K_m = O(\log(n))$. This proves the upper bound in Theorem 1 and concludes the construction of the algorithm. Before we conclude the proof, we finally discuss the computational complexity of the introduced algorithm to in detail.

\begin{figure*}%{0.48\textwidth}
        \centering
\begin{minipage}{0.91\textwidth}
    \begin{multicols}{2}
        \centering
        \begin{algorithmic}[1]
            \FOR{$t=1$ \TO $n$}
                \STATE \% Find the set of nodes containing $\vx[t]$
                \STATE $\kappa = \lambda$
                \STATE $\Set = \kappa$
                \WHILE{$s$ has children}
                    \STATE $\kappa = \kappa\nu$, where $\nu$ is the last letter of the child containing $\vx[t]$.
                    \STATE $\Set = \Set + \kappa$
                \ENDWHILE
                \STATE \% Check the index of the leaf node $\kappa$: if $\alpha_\kappa=0$, tree remains the same.
                \IF{$\alpha_\kappa=0$}
                    \STATE $\alpha_\kappa = \alpha_\kappa + 1$
                    \STATE $\T_\kappa = \T_\kappa + t$
                \STATE \% If $\alpha_\kappa=1$, create nodes $\kappa0$ and $\kappa1$.
                \ELSE
                    \STATE \% Train nodes $\kappa0$ and $\kappa1$.
                    \FORALL{$z \in \T_\kappa$}
                        \IF{$\vx[z] \in \R_{\kappa0}$}
                            \STATE $\T_{\kappa0} = \T_{\kappa0} + z$
                            \STATE $L_{\kappa0} = L_{\kappa0} \exp(-(d[z]-\vw_{\kappa0}^T\vx[z])^2/2a)$
                            \STATE $P_{\kappa0} = L_{\kappa0}$
                            \STATE $\vR_{\kappa0} = \vR_{\kappa0} + \vx[z] \vx[z]^T$
                            \STATE $\vw_{\kappa0} = \vw_{\kappa0} + \vR_{\kappa0} \backslash (\vx[z] (d[z]-\vw_{\kappa0}^T\vx[z]))$
                        \ELSE
                            \STATE \% Do the similar for node $\kappa1$.
                        \ENDIF
                    \ENDFOR
                    \FORALL{$\kappa \in \Set$}
                        \STATE $P_{\kappa} = (P_{\kappa0} P_{\kappa1} + L_{\kappa})/2$
                    \ENDFOR
                    \STATE \% Find the child containing $\vx[t]$ and perform tree updates.
                    \STATE $\nu = 0$, if $\vx[t] \in \R_{\kappa0}$, $\nu = 1$, otherwise.
                    \STATE $\kappa = \kappa\nu$
                    \STATE $\Set = \Set + \kappa$
                    \STATE $\alpha_{\kappa} = 1$
                \ENDIF
                \STATE \% Calculate combination weights and perform estimation.
                \FORALL{$\kappa_i \in \Pre(\kappa)$}
                    \IF{$\kappa_i = \lambda$}
                        \STATE $\pi_{\kappa_i} = 1/2$
                    \ELSIF{$\kappa_i \notin \{\lambda,\kappa\}$}
                        \STATE $\pi_{\kappa_i} = P_{\kappa_{i-1}\nu_i^c} \pi_{\kappa_{i-1}}/2$
                    \ELSE
                        \STATE $\pi_{\kappa_i} = P_{\kappa_{i-1}\nu_i^c} \pi_{\kappa_{i-1}}$
                    \ENDIF
                    \STATE $\mu_{\kappa_i} = \pi_{\kappa_i} L_{\kappa_i} / P_\lambda$
                    \STATE $\hat{d}_{\kappa_i} = \vw_{\kappa_i}^T \vx[t]$
                \ENDFOR
                \STATE $\hat{d} = \vmu^T \vec{\hat{d}}$
                \STATE $e = d[t] - \hat{d}$
                \STATE \% Perform algorithmic updates.
                \FORALL{$\kappa_i \in \Pre(\kappa)$}
                    \STATE $L_{\kappa_i} = L_{\kappa_i} \exp(-(d[t]-\hat{d}_{\kappa_i})^2/(2a))$
                    \IF{$\kappa_i = \kappa$}
                        \STATE $P_{\kappa_i} = L_{\kappa_i}$
                    \ELSE
                        \STATE $P_{\kappa_i} = (P_{\kappa_i0}P_{\kappa_i1}+L_{\kappa_i})/2$
                    \ENDIF
                    \STATE $R_{\kappa_i} = R_{\kappa_i} + \vx[t] \vx[t]^T$
                    \STATE $w_{\kappa_i} = w_{\kappa_i} + R_{\kappa_i} \backslash (\vx[t](d[t]-\hat{d}_{\kappa_i}))$
                \ENDFOR
            \ENDFOR
        \end{algorithmic}
    \end{multicols}
    \caption{The pseudocode of the Incremental Decision Tree (IDT) regressor}\label{alg}
    \end{minipage}
\end{figure*}

The computational complexity for the construction of the incremental decision tree is $O(|\Pre(\kappa)|)$, where $\kappa$ represents a leaf node of the incremental decision tree (see lines $2-35$ of the algorithm in Fig. \ref{alg} and note that $|\T_\kappa| \leq |\Pre(\kappa)|$). The computational complexity of the sequential weighting method is $O(|\Pre(\kappa)|)$ (see \eqref{eq:final_estimate} and lines $36-49$ of the algorithm in Fig. \ref{alg}). According to the incremental hierarchical partitioning method described, the number of light nodes on the tree (see Fig. \ref{fig:evolution}) is $t$ at time $t$, therefore we may observe a decision tree of depth $n$, i.e., $|\Pre(\kappa)|=n$, in the worst-case scenario, e.g., when $\vx[t] = [A,\dots,A]^T$ for all $t$. Hence, the computational complexity of the algorithm over a data length of $n$ is upper bounded by $O(n)$. Although theoretically the computational complexity of the algorithm is upper bounded by $O(n)$, in many real life applications the regressor vectors converge to stationary distributions \cite{sayed_book}. Hence, in such practical applications, the computational complexity of the algorithm can be upper bounded by $O(\log(n))$ as discussed in Remark 2. We emphasize that in order to achieve the computational complexity $O(\log(n))$, we do not require any statistical assumptions, instead it is sufficient that the regressor vectors are evenly (to some degree) distributed in the regressor space. This concludes the proof of Theorem 1. \hfill $\square$

{\bf Remark 1:}
Note that the algorithm in Fig. \ref{alg} achieves the performance of the best piecewise linear model having $O(\log(n))$ partitions with a regret of $O(p\log^2(n))$. In the most generic case, i.e., for an arbitrary piecewise model $m$ having $O(K_m)$ partitions, the introduced algorithm still achieves a regret of $O(pK_m\log(n/K_m))$. This indicates that for models having $O(n)$ partitions, the introduced algorithm achieves a regret of $O(pn)$, hence the performance of the piecewise model cannot be asymptotically achieved. However, we emphasize that no other algorithm can achieve a smaller regret than $O(pn)$ \cite{CTW}, i.e., the introduced algorithm is optimal in a strong minimax sense. Intuitively, this lower bound can be justified by considering the case, in which the regressor vector at time $t$ falls into the $t$th region of the piecewise model.

{\bf Remark 2:}
As mentioned in Remark 1 (and also can be observed in \eqref{eq:first}), no algorithm can converge to the performance of the piecewise linear models having $O(n)$ disjoint regions. Therefore, we can limit the maximum depth of the tree by $O(\log(t))$ at each time $t$ to achieve a low complexity implementation. With this limitation and according to the update rule of the tree, we can observe that while dividing a region into two disjoint regions, we may be forced to perform $O(t)$ computations due to the accumulated regressor vectors (since we no longer have $|\T_\kappa| \leq |\Pre(\kappa)|$ but instead have $|\T_\kappa| \leq t$). However, since a regressor vector is processed by at most $O(\log(n))$ nodes for any $n$, the average computational complexity of the update rule of the tree remains $O(\log(n))$. Furthermore, the performance of this low complexity implementation will be asymptotically the same as the exact implementation provided that the regressor vectors are evenly distributed in the regressor space, i.e., they are not gathered around a considerably small neighborhood. This result follows when we multiply the tree construction regret in \eqref{eq:second} by the total number of accumulated regressor vectors, whose order, according to the above condition, is upper bounded by $o(n/\log(n))$.

{\bf Remark 3:}
We emphasize that the node indexes, i.e., $\alpha_\kappa$'s, determines when to create finer regions. According to the described procedure, if a node at depth $l$ is partitioned into smaller regions, then its $i$th predecessor, i.e., $\kappa_i\in\Pre(\kappa)$, has observed at least $l-i$ different regressor vectors. Hence, a child node is created when coarser regions (i.e., predecessor nodes) are sufficiently trained. In this sense, we introduce new nodes to the tree according to the current status of the tree as well as the most recent data. We also point out that, in this paper, we divide each region from its midpoint (see Fig. \ref{fig:evolution}) to maintain universality. However, this process can also be performed in a data dependant manner, e.g., one can partition each region using the hyperplane that is perpendicular to the line joining two regressor vectors in that region. If there are more than two accumulated regressor vectors, then more advanced methods such as support vectors and anomaly detectors can be used to define a separator hyperplane. All these methods can be straightforwardly incorporated into our framework to produce different algorithms depending on the regression task.

%\begin{table*}[t]
%  %\centering
%  %\resizebox{.4\columnwidth}{!}{
%  \begin{tabular}{|l||*{6}{c|}} \hline
%  \backslashbox[3em]{Data Sets\;\;\;\; \kern-2em}{\kern-2em  \;\;\;\; Algorithms}
%  &\makebox{IDCTW}&\makebox{IDCTW-lc}&\makebox{CTW}&\makebox{LR}&\makebox{VF}&\makebox{FNF} \\\hline\hline
%  \hspace{.35cm} Lorenz Attractor      & $0.0354$ & $0.0354$ & $0.0360$ & $0.0422$ & $0.0371$ & $0.0403$ \\\hline
%                 Mackey-Glass Sequence & $0.0244$ & $0.0244$ & $0.0291$ & $0.1031$ & $0.0354$ & $0.0373$ \\\hline
%  \hspace{.45cm} Chua's Circuit        & $0.0266$ & $0.0266$ & $0.0316$ & $0.0372$ & $0.0372$ & $0.0802$ \\\hline
%  \hspace{.50cm} Duffing Map           & $0.0124$ & $0.0124$ & $0.0933$ & $0.2200$ & $0.2308$ & $0.5977$ \\\hline
%  \hspace{.30cm} R\"{o}ssler Attractor & $0.1468$ & $0.1468$ & $0.1798$ & $0.1983$ & $0.1765$ & $0.1580$ \\\hline
%  \hspace{.35cm} Tinkerbell Map        & $0.0834$ & $0.0834$ & $0.1068$ & $0.1083$ & $0.1034$ & $0.0885$ \\\hline
%  \end{tabular}
%  %}
%  \caption{Time accumulated normalized errors of the proposed algorithms. Each dimension of the data sets is normalized between $[-1,1]$.}\label{tab:error_table}
%\end{table*}

\subsection{Proof of Theorem 2}
We begin our proof by emphasizing that the introduced algorithm converges to the best linear model in each region with a regret of $O(p\log^2(n))$ for any finite regression parameter $\vv_m$ (since $\norm{\vv_m}\leq\delta Gp\log(n)$) as already proven in Theorem 1. Therefore, using any other linear model yields a higher regret. Hence, say we define a suboptimal affine model by applying Taylor's theorem to a twice differentiable function $f \in \F$ about the midpoint of each region. Let $\hat{d}_s[t]$ denote the prediction of this suboptimal affine regressor. Then, we have
\[
\sum_{t=1}^n \left( d[t] - \hat{d}[t] \right)^2 \leq \sum_{t=1}^n \left( d[t] - \hat{d}_s[t] \right)^2 + O(p\log^2(n)).
\]
Now applying the mean value theorem with the Lagrange form of the remainder, we obtain
\begin{align}
  & \sum_{t=1}^n \left( d[t] - \hat{d}_s[t] \right)^2 - \sum_{t=1}^n \left( d[t] - \hat{d}_f[t] \right)^2 \nn\\
  & \leq 2A \sum_{t=1}^n \Bigg\{ \sum_{i=1}^p \sum_{j=1}^p \frac{\partial^2 f(\vx)}{\partial x_i \partial x_j} \bigg{|}_{\vx=\vb} (x_i[t]-a_{\kappa,i})(x_j[t]-a_{\kappa,j}) \Bigg\}, \nn
\end{align}
for some $m \in \M_n'$ and $\vb \in \R_\kappa$, where $\va_\kappa \triangleq [a_{\kappa,1},\dots,a_{\kappa,p}]^T$ is the midpoint of the region $\R_\kappa$. Maximizing this upper bound with respect to $\vx$ we obtain
\begin{align}
  & \sum_{t=1}^n \left( d[t] - \hat{d}[t] \right)^2 - \sum_{t=1}^n \left( d[t] - \hat{d}_f[t] \right)^2 \nn\\
    & \hspace{2.5cm} \leq 2A D p^2 n \frac{A^2}{O(\log^{2/p}(n))} + O(p\log^2(n)) \nn\\
    & \hspace{2.5cm} \leq o(p^2n). \nn
\end{align}
This concludes the proof of Theorem 2. \hfill $\square$

\section{Simulations}\label{sec:Simulations}
In this section, we investigate the performance of the introduced algorithm with respect to various methods under several benchmark scenarios. Throughout the experiments, we denote the incremental decision tree algorithm of Theorem 1 by ``IDT'', the context tree weighting algorithm of \cite{CTW} by ``CTW'', the linear regressor by ``LR'', the Volterra series regressor by ``VSR'' \cite{volterra}, the sliding window Multivariate Adaptive Regression Splines of \cite{mars1,mars2} by ``MARS'', and the Fourier nonlinear regressor of \cite{fnf} by ``FNR''. The combination weights of the LR, VSR, and FNR are updated using the recursive least squares (RLS) algorithm \cite{sayed_book}. Unless otherwise stated, the CTW algorithm has depth $2$, the VSR, FNR, and MARS algorithms are second order, and the MARS algorithm uses $21$ knots with a window length of $500$ that shifts in every $200$ samples.

\begin{table}[t]
  \centering
  %\resizebox{.75\columnwidth}{!}{
  \begin{tabular}{| c | c |}
    \hline
    Algorithm & Computational Complexity \\ \hline
    IDT       & $O\left( p^2\log(n) \right)$ \\ \hline
    CTW       & $O\left( p^2d \right)$ \\ \hline
    LR        & $O\left( p^2 \right)$ \\ \hline
    VSR       & $O\left( p^{2r} \right)$ \\ \hline
    MARS      & $O\left( rbw^3 \right)$ \\ \hline
    FNR       & $O\left( (pr)^{2r} \right)$ \\ \hline
  \end{tabular}
  %}
  \caption{Comparison of the computational complexities of the proposed algorithms with the corresponding update rules. In the table, $p$ represents the dimensionality of the regressor space, $d$ represents the depth of the trees in the respective algorithms, and $r$ represents the order of the corresponding filters and algorithms. For the MARS algorithm (particularly, the fast MARS algorithm, cf. \cite{mars2}), $b$ represents the number of basis functions and $w$ represents the window length.}\label{tab:complexity}
\end{table}

In Table \ref{tab:complexity}, we provide the computational complexities of the proposed algorithms. We emphasize that although the computational complexity to create and run the incremental decision tree is $O(\log(n))$, the overall computational complexity of the algorithm is $O(p^2\log(n))$ due to the universal linear regressors at each region. Particularly, since the universal linear regressor at each region has a computational complexity of $O(p^2)$, the overall computational complexity of $O(p^2\log(n))$ follows. However, this universal linear regressor can be straightforwardly replaced with any linear (or nonlinear) regressor in the literature. For example, if we use the LMS algorithm to update the parameters of the linear regressor instead of using the universal algorithm for this update, the computational complexity of the overall structure becomes $O(p\log(n))$. Hence, although the computational complexity of the original IDT algorithm is $O(\log(n))$, this computational complexity may increase according to the computational complexity of the node regressors.

In this section, we first illustrate the performances of the proposed algorithms for a synthetic piecewise linear model that do not match the modeling structure of any of the above algorithms. We then consider the prediction of chaotic signals (generated from Duffing and Tinkerbell maps) and well-known data sequences such as Mackey-Glass sequence and Chua's circuit \cite{Hero}. Finally, we consider the prediction of real life examples that can be found in various benchmark data set repositories such as \cite{uci,delve}.

\subsection{Synthetic Data}
In this subsection, we consider the scenario where the desired data is generated by the following piecewise linear model
\begin{equation}\label{eq:synthetic}
 \resizebox{\hsize}{!}{$
  d[t] =
  \begin{cases}
  x_1[t]+x_2[t]+n[t] & \text{, if $\norm{\vx[t]}^2 \in [0,0.1]\cup[0.5,1]$ }\\
  -x_1[t]-x_2[t]+n[t] & \text{, otherwise}
  \end{cases},
  $}
\end{equation}
and $\vx[t] = [x_1[t], x_2[t]]^T$ are sample functions of a jointly Gaussian process of mean $[0,0]^T$ and covariance matrix $\vI$, and $n[t]$ is a sample function from a zero mean white Gaussian process with variance $0.1$. Note that the piecewise model in \eqref{eq:synthetic} has circular regions, which cannot be represented by hyperplanes or twice differentiable functions. Hence, the underlying relationship between the desired data and the regressor vectors cannot be exactly modeled using any of the proposed algorithms.

\begin{figure}[t]
  \centering
  \includegraphics[width=0.48\textwidth]{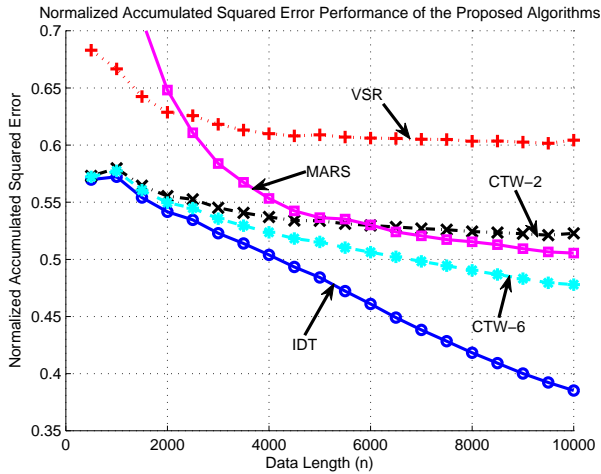}\\
  \caption{Normalized accumulated squared error performances for the piecewise linear model in \eqref{eq:synthetic} averaged over $10$ trials.}\label{fig:synthetic}
\end{figure}

In Fig. \ref{fig:synthetic}, we present the normalized accumulated squared errors of the proposed algorithms averaged over $10$ trials. For this experiment, ``CTW-2'' and ``CTW-6'' show the performances of the CTW algorithm with depths $2$ and $6$, respectively. Since the performances of the LR and FNR algorithms are incomparable with the rest of the algorithms, they are not included in the figure for this experiment. Fig. \ref{fig:synthetic} illustrates that even for a highly nonlinear system \eqref{eq:synthetic}, our algorithm significantly outperforms the other algorithms. The normalized accumulated error of the introduced algorithm goes to the variance of the noise signal as $n$ increases, unlike the rest of the algorithms, whose performances converge to the performance of their optimal batch variants as $n$ increases. This observation can be seen in Fig. \ref{fig:synthetic}, where the normalized cumulative error of the IDT algorithm steadily decreases since the IDT algorithm creates finer regions as the observed data length increases. Hence, even for a highly nonlinear model such as the circular piecewise linear model in \eqref{eq:synthetic}, which cannot be represented via hyperplanes, the IDT algorithm can well approximate this highly nonlinear relationship by incrementally introducing finer partitions as the observed data length increases.

\begin{figure}[t]
  \centering
  \includegraphics[width=0.48\textwidth]{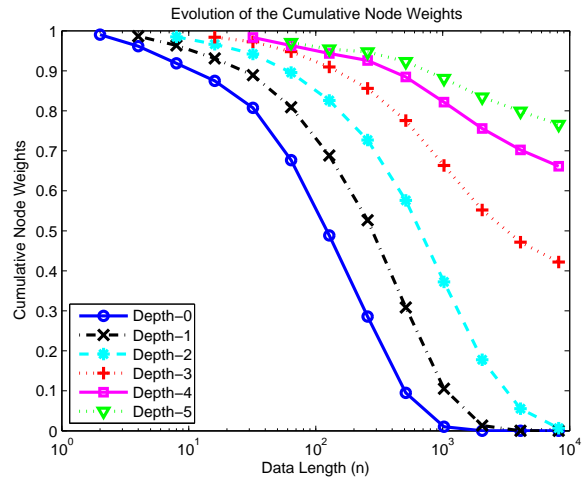}\\
  \caption{Evolution of the normalized cumulative node weights at the corresponding depths of the tree for the piecewise linear model in \eqref{eq:synthetic} averaged over $10$ trials.}\label{fig:weights}
\end{figure}

Furthermore, even though the depth of the introduced algorithm is comparable with the CTW-6 algorithm over short data sequences, the performance of our algorithm is superior with respect to the CTW-6 algorithm. This results since the IDT algorithm intrinsically eliminates the extremely finer models at the early processing stages and introduces them whenever they are needed, unlike the CTW-6 algorithm. This procedure can be observed in Fig. \ref{fig:weights}, where the IDT algorithm introduces finer regions (i.e., nodes with higher depths) to the hierarchical model as the coarser regions becomes unsatisfactory. Since the universal algorithms such as CTW distribute a ``budget'' into numerous experts, as the number of experts increases, the performance of such algorithms deteriorate. On the other hand, the introduced algorithm intrinsically limits the number of experts according to the unknown data length at each iteration, hence we avoid such possible performance degradations as can be observed in Fig. \ref{fig:weights}.

\begin{figure*}[t]
    \centering
    \begin{subfigure}[b]{0.48\textwidth}
        \centering
        \includegraphics[width=\textwidth]{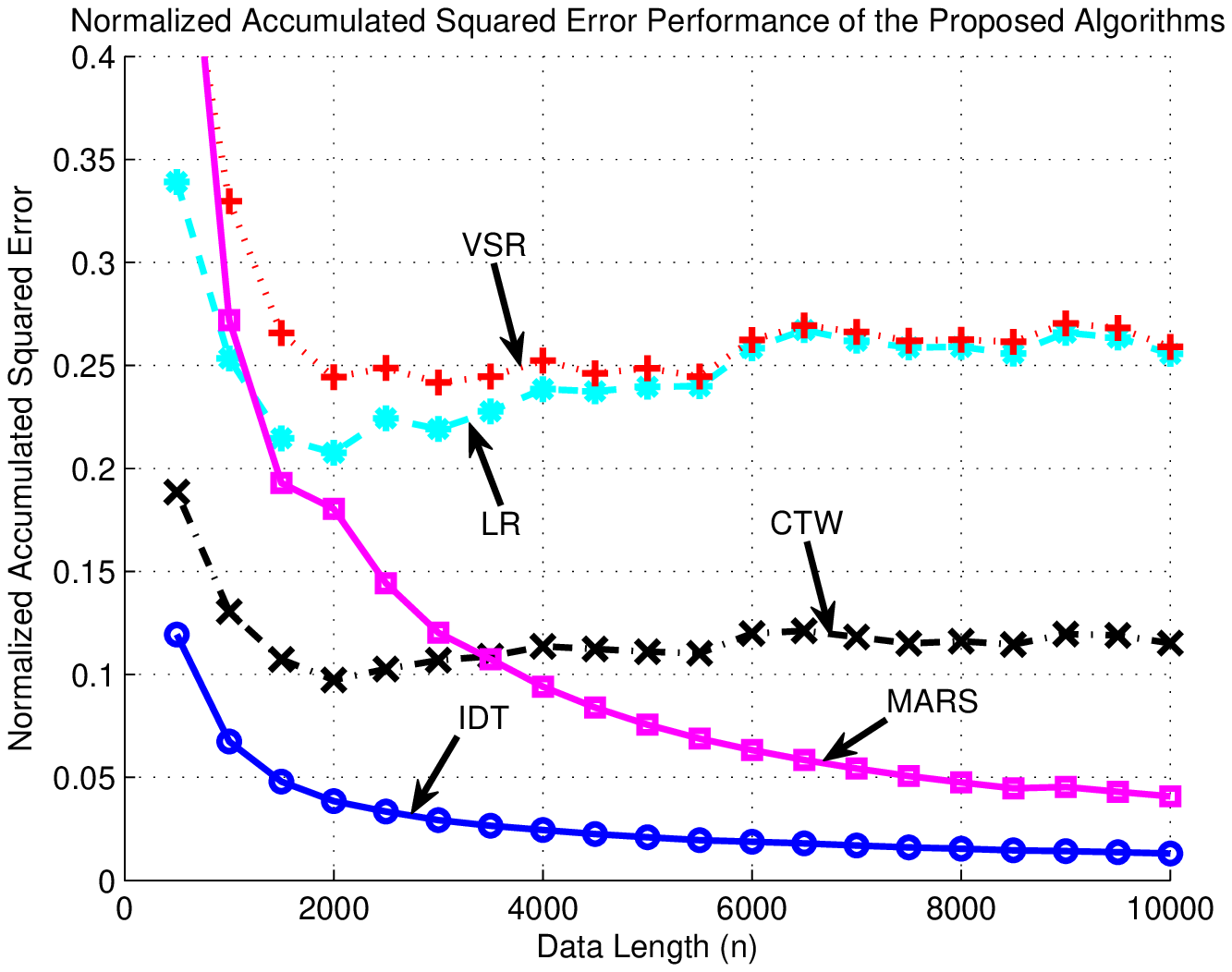}\\
        \caption{}\label{fig:duffing}
    \end{subfigure}
    \begin{subfigure}[b]{0.48\textwidth}
        \centering
        \includegraphics[width=\textwidth]{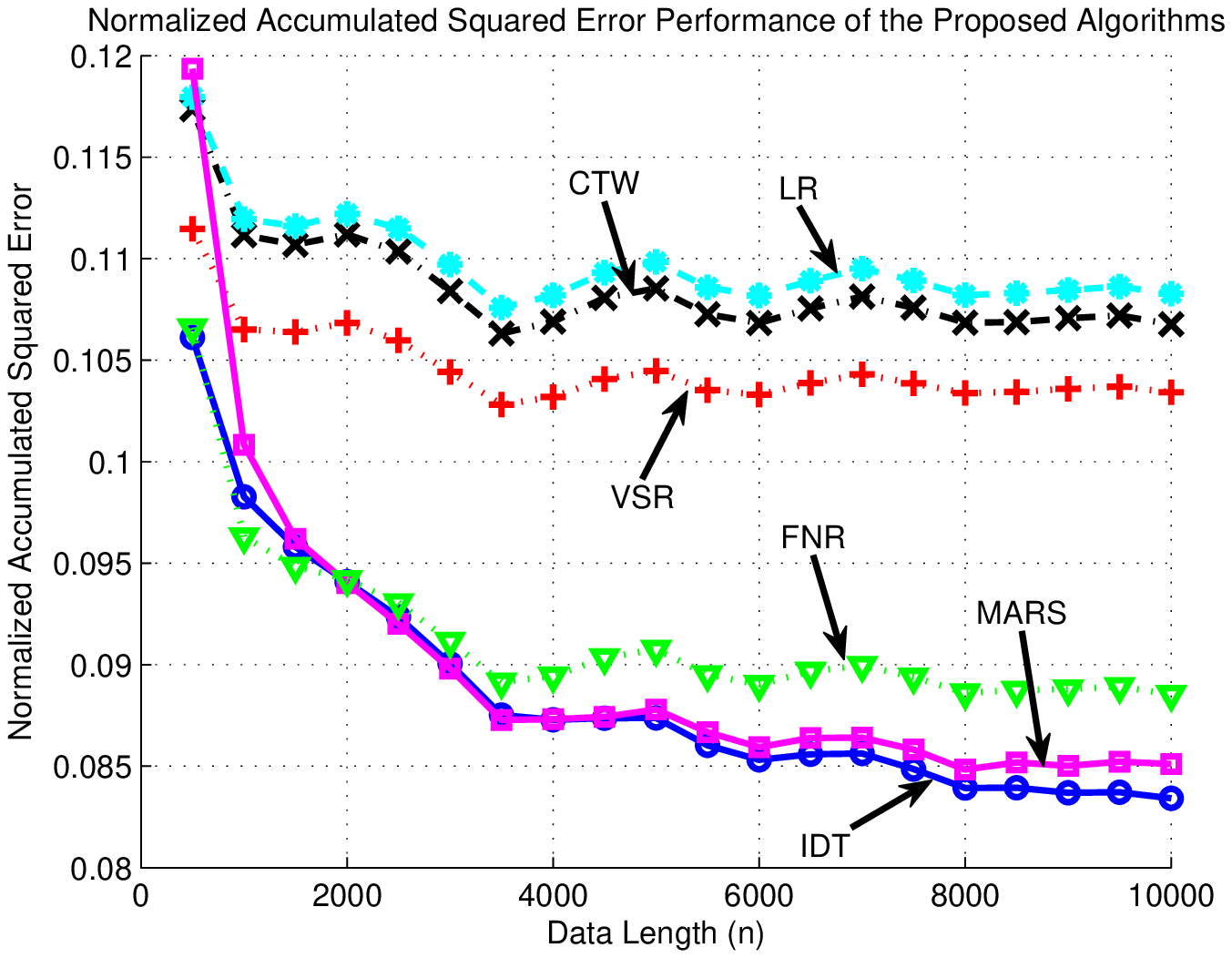}\\
        \caption{}\label{fig:tinkerbell}
    \end{subfigure}
    \caption{Normalized accumulated squared error performances for the chaotic data generated by (a) the Duffing map in \eqref{eq:duffing} (b) the Tinkerbell map in \eqref{eq:tinkerbell1} and \eqref{eq:tinkerbell2}.}\label{fig:chaotic}
\end{figure*}

\subsection{Chaotic Data}
In this subsection, we consider prediction of the chaotic signals generated from the Duffing and Tinkerbell maps. The Duffing map is generated by the following discrete time equation
\begin{equation}\label{eq:duffing}
  x[t+1] = a x[t] - (x[t])^3 - b x[t-1],
\end{equation}
where we set $a=2.75$ and $b=0.2$ to produce the chaotic behavior \cite{CTW2,chaos}. The Tinkerbell map is generated by the following discrete time equations
\begin{align}
  x[t+1] &= (x[t])^2 - (y[t])^2 + ax[t] + by[t] \label{eq:tinkerbell1} \\
  y[t+1] &= 2x[t]y[t] + cx[t] + dy[t], \label{eq:tinkerbell2}
\end{align}
where we set $a=0.9$, $b=-0.6013$, $c=2$, and $d=0.5$ \cite{CTW,chaos}. We emphasize that these values are selected to generate the well-known chaotic behaviors of these attractors.

Fig. \ref{fig:duffing} and Fig. \ref{fig:tinkerbell} shows the normalized accumulated squared error performances of the proposed algorithms. We emphasize that due to the chaotic nature of the signals, we observe non-uniform curves in Fig. \ref{fig:chaotic}. Since the conventional nonlinear and piecewise linear regression algorithms commit to a priori partitioning and/or basis functions, their performances are limited by the performances of the optimal batch regressors using these prior partitioning and/or basis functions as can be observed in Fig. \ref{fig:chaotic}. Hence, such prior selections result in fundamental performance limitations for these algorithms. For example, in the CTW algorithm, the partitioning of the regressor space is set before the processing starts. If this partitioning does not match with the underlying partitioning of the regressor space, then the performance of the CTW algorithm becomes highly unsatisfactory as seen in Fig. \ref{fig:chaotic}. On the other hand, the introduced algorithm illustrates a robust and superior performance while learning the underlying chaotic relationships, whereas the rest of the algorithms yields an inconsistent performance due to the chaotic nature of the signals and the limited modeling power of these algorithms.

\subsection{Benchmark Sequences}
In this subsection, we consider the prediction of the Mackey-Glass and Chua's circuit sequences. The Mackey-Glass sequence is defined by the following differential equation
\begin{equation}\label{eq:mackeyglass}
  \frac{dx[t]}{dt} = \frac{\beta x[t-\tau]}{1+(x[t-\tau])^n} - \gamma x[t],
\end{equation}
where we set $\beta=2$, $\gamma=1$, $\tau=2$, and $n=10$ with the initial condition $x[t]=0.5$ for $t<0$. We generate the time series using the fourth order Runge-Kutta method. The Chua's circuit is generated according to the following differential equations
\begin{equation}\label{eq:chua}
  \frac{dx}{dt} = \alpha(y-x-f(x)), \,\, \frac{dy}{dt} = x-y+z, \,\, \frac{dz}{dt} = -\beta y,
\end{equation}
where we dropped the time index for notational simplicity, and $f(x) = m_1 x + 0.5(m_0-m_1)(|x+1|-|x-1|)$, $\alpha = 15.6$, $\beta=28$, $m_0 = -1.143$, $m_1 = -0.714$ with initial conditions $[x,y,z] = [0.7,0,0]$ for $t<0$.

\begin{figure*}[t]
    \centering
    \begin{subfigure}[b]{0.48\textwidth}
        \centering
        \includegraphics[width=\textwidth]{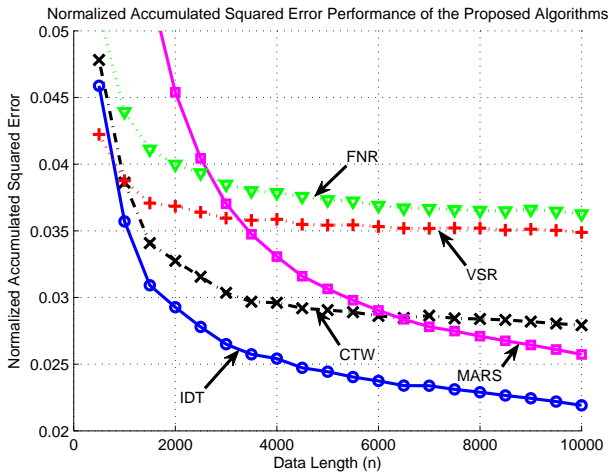}\\
        \caption{}\label{fig:mackeyglass}
    \end{subfigure}
    \begin{subfigure}[b]{0.48\textwidth}
        \centering
        \includegraphics[width=\textwidth]{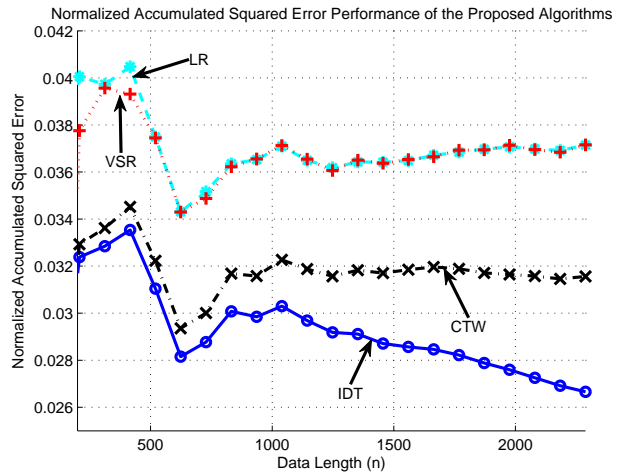}\\
        \caption{}\label{fig:chua}
    \end{subfigure}
    \caption{Normalized accumulated squared error performances for (a) the Mackey-Glass sequence in \eqref{eq:mackeyglass} (b) the Chua's circuit sequence in \eqref{eq:chua}.}\label{fig:benchmark}
\end{figure*}

Fig. \ref{fig:mackeyglass} and Fig. \ref{fig:chua} shows the normalized accumulated squared error performances of the proposed algorithms for the Mackey-Glass and Chua's circuit sequences, respectively. We emphasize that due to the chaotic nature of the signals, we observe non-uniform curves in Fig. \ref{fig:benchmark}. In the figures, the algorithms with incomparable (i.e., unsatisfactory) performance are omitted. Fig. \ref{fig:benchmark} presents that the IDT algorithm achieves an average of $20\%$ relative gain in the performance with respect to the other algorithm and can accurately predict these well-known data sequences.

\subsection{Real Data}
In this subsection, we evaluate the performance of the proposed algorithms for two well-known real data sets in machine learning literature, namely ``kinematics'' and ``pumadyn'' \cite{uci,delve}. The kinematics data set involves a realistic simulation of the forward dynamics of an $8$ link all-revolute robot arm and the task is to predict the distance of the end-effector from a target. Among its variants, we used the one having $9$ attributes and being nonlinear as well as medium noisy. The pumadyn data set involves a realistic simulation of the dynamics of a Puma $560$ robot arm. The task in these datasets is to predict the angular acceleration of one of the robot arm's links. Among its variants, we used the one having $9$ attributes and being nonlinear as well as medium noisy.

\begin{figure*}[t]
    \centering
    \begin{subfigure}[b]{0.48\textwidth}
        \centering
        \includegraphics[width=\textwidth]{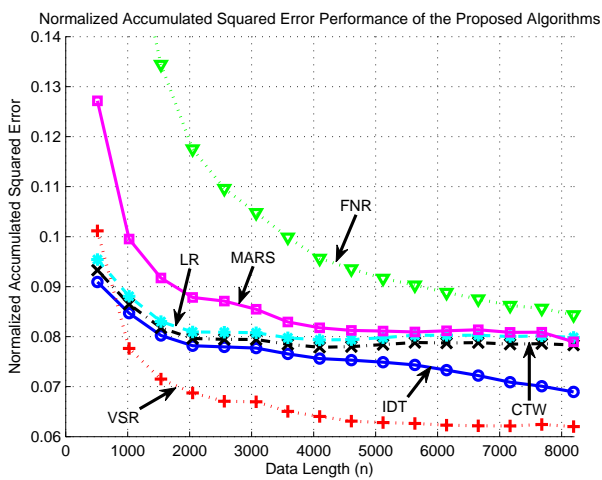}\\
        \caption{}\label{fig:kinematics}
    \end{subfigure}
    \begin{subfigure}[b]{0.48\textwidth}
        \centering
        \includegraphics[width=\textwidth]{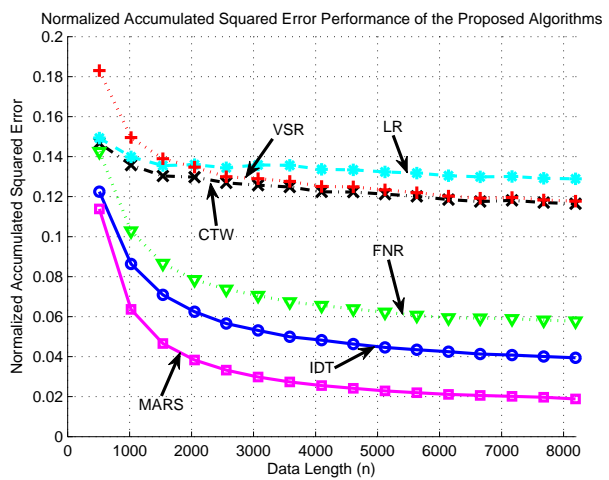}\\
        \caption{}\label{fig:pumadyn}
    \end{subfigure}
    \caption{Normalized accumulated squared error performances for (a) the ``kinematics'' (b) the ``pumadyn'' data sets.}\label{fig:real}
\end{figure*}

Fig. \ref{fig:kinematics} and Fig. \ref{fig:pumadyn} shows the normalized accumulated squared error performances of the proposed algorithms for the kinematics and pumadyn data sets, respectively. In the experiments, all dimensions of the regressor vector and desired data are normalized between $[-1,1]$. Although in Fig. \ref{fig:kinematics}, the VSR algorithm provides the best performance and in Fig. \ref{fig:pumadyn}, the MARS algorithm achieves the minimum accumulated error, the performances of these algorithms in the reciprocal experiments are highly unsatisfactory. This result implies that the data in the first experiment can be well approximated by Volterra series, whereas the model that generates the data in the second experiment is more inline with B-splines. Hence, the performances of these algorithms are extremely sensitive to the underlying structure that generates the data. On the other hand, the IDT algorithm nearly achieves the performance of the best algorithm in both experiments and presents a desirable performance under different scenarios. This result implies that the introduced algorithm can be used in various frameworks without any significant performance degradations owing to its guaranteed performance upper bounds without any statistical or structural assumptions.

\section{Concluding Remarks}\label{sec:Conclusion}
We study nonlinear regression of deterministic signals using an incremental decision tree, where the regressor space is partitioned using a nested structure and independent regressors are assigned to each region. In this framework, we introduce a tree based algorithm that sequentially increases its nonlinear modeling power and achieves the performance of the optimal twice differentiable function as well as the performance of the best piecewise linear model defined on the incremental decision tree. Furthermore, this performance is achieved only with a computational complexity logarithmic in the data length $n$, i.e., $O(\log(n))$ (under regularity conditions). We demonstrate the superior performance of the introduced algorithm over a series of well-known benchmark applications in the regression literature.

\bibliographystyle{IEEEtran}
\bibliography{my_references}

\end{document}